\documentclass{article} 
\usepackage{iclr2025_conference,times}

\usepackage{amsmath,amsfonts,bm}

\def\eqref#1{equation~\ref{#1}}

\def\1{\bm{1}}

\DeclareMathAlphabet{\mathsfit}{\encodingdefault}{\sfdefault}{m}{sl}
\SetMathAlphabet{\mathsfit}{bold}{\encodingdefault}{\sfdefault}{bx}{n}

\usepackage{hyperref}
\usepackage{url}

\title{FishNet++: Analyzing the capabilities of Multimodal Large Language Models in marine biology}
\iclrfinalcopy

\author{Faizan Farooq Khan$^1$,   \, Yousef Radwan$^1$, \, 
Eslam Abdelrahman$^1$, \, Abdulwahab Felemban$^1$ \\ 
\textbf{Aymen Mir}$^3$, \, \textbf{Nico K. Michiels}$^3$, \, \textbf{Andrew J. Temple}$^1$$^2$, \, 
\textbf{Michael L. Berumen}$^1$$^2$, \, \textbf{Mohamed Elhoseiny}$^1$ \\
$^1$King Abdullah University of Science and Technology\\
$^2$ Red Sea Research Center, KAUST \\
$^3$ Tübingen University \\
\footnotesize
}

\newcommand{\ours}{FishNet++\xspace}

\newcommand{\oldten}{495\xspace}
\newcommand{\newspe}{17,393\xspace}
\newcommand{\newten}{804\xspace}
\newcommand{\totalspe}{35,133\xspace}

\newcommand{\newcnt}{5,024\xspace}
\newcommand{\totcnt}{99,556\xspace}

\newcommand{\kp}{706,426\xspace}

\newcommand{\bbox}{119,399\xspace}

\definecolor{blond}{rgb}{0.98, 0.94, 0.75}
\usepackage[capitalize,noabbrev]{cleveref}
\usepackage{pifont}
\newcommand{\cmark}{\ding{51}}%
\newcommand{\xmark}{\ding{55}}%
\usepackage{multirow}
\usepackage{cleveref}

\usepackage{booktabs}
\usepackage{cleveref}
\usepackage{xspace}

\usepackage{amsmath}
\usepackage{pgfplots}

\usepackage{adjustbox}
\usepackage{wrapfig}
\usepackage{pgfplotstable}
\usepackage{array}
\usepackage{nicefrac}
\usepackage{subcaption}
\usepackage{wrapfig}
\usetikzlibrary{calc}
\usepackage[table]{xcolor}
\usepackage{multirow}
\usepackage{pifont}

\usepackage[noend]{algpseudocode}
\usepackage{algorithm}

\definecolor{blond}{rgb}{0.98, 0.94, 0.75}

\begin{document}

\maketitle

\begin{abstract}
Multimodal large language models (MLLMs) have demonstrated impressive cross-domain capabilities, yet their proficiency in specialized scientific fields like marine biology remains underexplored. In this work, we systematically evaluate state-of-the-art MLLMs and reveal significant limitations in their ability to perform fine-grained recognition of fish species, with the best open-source models achieving less than 10\% accuracy. This task is critical for monitoring marine ecosystems under anthropogenic pressure. To address this gap and investigate whether these failures stem from a lack of domain knowledge, we introduce FishNet++, a large-scale, multimodal benchmark. FishNet++ significantly extends existing resources with 35,133 textual descriptions for multimodal learning, 706,426 key-point annotations for morphological studies,  and 119,399 bounding boxes for detection. By providing this comprehensive suite of annotations, our work facilitates the development and evaluation of specialized vision-language models capable of advancing aquatic science. 
\end{abstract}
\section{Introduction}
\label{sec:intro}
Healthy aquatic ecosystems and the services they provide are essential for human survival~\cite{bioimp1, Basurto2025, BARBIER2017R507}. The health of these ecosystems and the volume and quality of ecosystem services are closely tied to changes in their biodiversity~\cite{1132294, Tett2013}. At a time when aquatic ecosystems are under intense threat from human activities such as fisheries, climate change, coastal development, and pollution, conservation and management interventions are critical in preserving and restoring ecosystem health. Most conservation efforts begin with basic documentation, recognition, and monitoring of biodiversity; in aquatic ecosystems, these efforts are complicated by their often remote and relatively inaccessible nature. As a result, they become time and labor-intensive processes that require expert knowledge to undertake what might otherwise be considered relatively menial tasks. When extrapolated to the global scale, this first step presents a critical bottleneck in our ability to generate the information required to make informed decisions and to take the essential conservation and management actions required to preserve the health of aquatic ecosystems. 

Recent advances in Multimodal Large Language Models (MLLMs) offer promising potential for automation across a variety of tasks, having demonstrated exceptional generalist skills in vision-language tasks~\cite{visualgpt, flamingo, flava, llava, minigpt, minigptv2}. However, it is unclear if this proficiency translates to the fine-grained, expert-level knowledge required for marine species recognition to support conservation efforts. 

To address this, we conduct a systematic analysis to answer a crucial question: Do state-of-the-art Multimodal Large Language Models (MLLMs) possess the specialized knowledge required for aiding marine ecology conservation efforts, or do their capabilities degrade when confronted with fine-grained, out-of-distribution data? We first probe the recognition level of leading MLLMs by evaluating their zero-shot species recognition performance, revealing that even the most capable models lack domain knowledge. Qwen2.5-VL achieves just 6.2\% accuracy on frequent species and 0.2\% on rare species. 

This initial finding motivates a deeper diagnostic question: Does this failure stem from a core lack of domain-specific knowledge or from inadequate visual perception of fine-grained features in the marine domain? To disentangle these factors, we design three targeted tasks. 1) Domain Knowledge: We assess the models' domain knowledge by evaluating their ability to relate common names to scientific names and vice versa. 2) Visual Domain Knowledge: We evaluate the visual domain knowledge by testing whether the models can verify the presence/absence of a given species in an image. 3) Perception Capabilities: We test how well the models can (a) locate the species with a bounding box, and (b) pinpoint specific morphological structures through key-part localization.

To facilitate this investigation, we introduce \ours, a large-scale, multi-modal benchmark designed not only to diagnose these limitations but also to help improve recognition. \ours comprises \totcnt images across \newspe fish species, enriched with \kp key-point annotations, \bbox bounding boxes, and detailed textual descriptions. We leverage this benchmark to first quantify the zero-shot performance of MLLMs and then show their lack of domain knowledge and how improvements can be achieved.

To summarize, our contributions are:
\begin{itemize}
\item We conduct the first large-scale analysis of MLLMs in the marine domain, revealing critical performance gaps in their zero-shot knowledge of marine species.
\item We conduct a detailed diagnostic analysis, deconstructing this poor performance across three tasks to disentangle failures in semantic knowledge from visual perception.
\item We introduce \ours, a comprehensive multi-modal benchmark with annotations for open-vocabulary recognition, detection, and keypoint localization, serving both as a diagnostic tool for evaluating MLLMs and as a resource for developing stronger marine domain-aware models.
\item We demonstrate that the identified knowledge gap can be mitigated, showing that fine-tuning on \ours substantially boosts MLLM performance.
\end{itemize}

\section{Related Work}
\label{sec:related}

\paragraph{Open-Vocabulary Recognition.}
The task of open-vocabulary recognition has evolved from early works like~\cite{ogovc}, which introduced joint image–word embeddings for semantic segmentation, allowing models to go beyond fixed label sets. This line of research gained momentum with the advent of large-scale pretrained models such as BERT~\cite{bert} for text and CLIP~\cite{clip}, which aligned vision and language embeddings for zero-shot classification. CLIP’s success led to extensions for open-vocabulary detection~\cite{vild}, segmentation~\cite{lseg}, and classification~\cite{ocls, anytime}. While CLIP-like models~\cite{clip, openclip, siglip} perform well in general settings, they remain suboptimal in fine-grained, open-world recognition, likely due to limited taxonomic understanding and dataset bias. This is discussed further in~\cref{sub_ovc}.

\paragraph{Dense Recognition Tasks.} 
Classical dense recognition methods rely on bounding-box or pixel-wise prediction. One-stage detectors like YOLO~\cite{redmon2016lookonceunifiedrealtime} unify localization and classification for real-time inference (up to 155 fps). Two-stage detectors such as Faster R-CNN~\cite{ren2016fasterrcnn} generate region proposals before classification, and Mask R-CNN~\cite{he2018maskrcnn} extends this by adding a segmentation branch. Transformer-based DETR~\cite{carion2020endtoendobjectdetectiontransformers} reframes detection as set prediction using an encoder–decoder transformer, removing the need for non-maximum suppression and anchors. For segmentation, models like FCN~\cite{long2015fcn}, DeepLab~\cite{chen2017deeplab}, MaskFormer~\cite{maskformer}, and SAM~\cite{kirillov2023segment} demonstrate strong generalization. In fish imagery, these architectures (e.g., YOLO, Mask R-CNN) are widely applied with domain-specific tuning. Given its efficiency, we adopt YOLO-based~\cite{redmon2016lookonceunifiedrealtime} models for our dense-prediction tasks.

\paragraph{Species Recognition.} 
Fine-grained species recognition is a major focus in ecology and biodiversity monitoring, which poses unique challenges (e.g., high intra-class variance, inter-class similarity, and class imbalance)~\cite{fish4know, seampd21, fishabun, imgdata, ncfm, inat, zhuang2020wildfish++}. For aquatic environments specifically, new datasets have been released. These include Fishnet Open Images Database~\cite{kay2021fishnet}, an open images dataset of $86,000$ of fish from $34$ species taken from vessel-bourne cameras, which highlights domain conditions like murky water, skewed species distribution, and occlusion. AutoFish~\cite{bengtson2025autofishdatasetbenchmarkfinegrained}, another dataset with $1,500$ controlled-setup collected images of $454$ fish instances annotated with segmentation and IDs. We compare \ours dataset with further existing datasets in~\cref{tab1}.

\begin{table}[!t]
\caption{Comparison with existing datasets for fish recognition tasks. FishNet++ provides textual descriptions for more than 35,000 species, while previous datasets only provide species labels. FishNet++ supports additional tasks for detection, key-part localization, and segmentation.}
\scalebox{0.8}{
\footnotesize
\begin{tabular}{l|ccc |ccc}
\toprule
\multirow{2}{*}{Datasets} & \multicolumn{3}{c}{Properties} & \multicolumn{3}{c}{Tasks} \\
 & Images & Species & Descriptions & Open-Vocabulary & Detection & Part-Location\\
\midrule
Fish4-Knowledge-A~\cite{fish4know} & 27,370 & 23 & 0 &\xmark & \cmark  & \xmark \\
SEAMPD21~\cite{seampd21} & 28,328 & 130 & 0 & \xmark & \xmark & \xmark    \\
Fish-gres~\cite{fishgres} & 3,248 & 8 & 0 & \xmark & \xmark & \xmark   \\
Mediterranean Fish Species~\cite{mfs} & $\approx$40,000 & 20 & 0 & \xmark & \xmark & \xmark \\
Fish Abundance~\cite{fishabun} & 4,909 & 50  & 0 & \xmark & \xmark & \xmark \\
Image Dataset~\cite{imgdata} & 33,805 & 30 & 0 & \xmark & \xmark & \xmark  \\
NCFM~\cite{ncfm} & 16,915 & 8 & 0 & \xmark & \xmark & \xmark  \\
iNaturalist\_{Fish}~\cite{iNaturalist} & 54,006 & 369  & 0 & \xmark & \xmark & \xmark  \\
WildFish++~\cite{zhuang2020wildfish++} & 103,034 & 2,348 & 0 & \xmark & \xmark & \xmark  \\
FishNet~\cite{fishnet} & 94,532 & 17,357  & 0 & \xmark & \cmark & \xmark \\
\textbf{Ours} & 99,556 & 17,393 & 35,133 & \cmark & \cmark & \cmark   \\
\midrule
\end{tabular}
}


\vspace{-2mm}
    \label{tab1}
\end{table}

\paragraph{MultiModal Large Language Models (MLLMs).} MLLMs have advanced multimodal understanding and reasoning through large-scale pretraining, supervised fine-tuning, and often RLHF~\cite{gpt4o, gpt4v, gemini, deepseek, qwen25, mixtralexperts, llama, qwen2, rlhf}. Scaling models and data has been key to their success, yet they still struggle with long or complex contexts~\cite{Yin_2024}. To address this, retrieval-augmented generation(RAG)~\cite{lewis,izacard} approaches have emerged as a practical solution, enabling models to access and reason over extended external information while reducing hallucinations and improving factual grounding. Recent works like~\cite{hal1, asai2023selfrag} extend RAG to long-form reasoning, multi-hop QA, and vision-centric tasks, e.g., MuRAG~\cite{murag} with image-text memory banks. In this work, we also show RAGs as a potential approach to enhance the performance of MLLMs for the open-vocabulary recognition task.

\section{\ours}
\label{sec:data}
While it is estimated that over 95\% of the world's bird species have been described~\cite{barrowclough2016}, the vast majority of marine life remains a mystery, with some estimates suggesting over 90\% of species are yet to be discovered~\cite{mora2011}. Despite this enormous knowledge gap, the focus of the computer vision community has predominantly been on terrestrial animals~\cite{cub, birdsnap, nabirds}. To help bridge this disparity and advance aquatic science, we introduce \ours, a large-scale, multi-modal benchmark developed from the original FishNet dataset~\cite{fishnet}. Our primary goal is to enable the development of models capable of large-scale, language-based species recognition, a foundational step towards the ultimate challenge of identifying unseen or newly discovered species.

\ours is enriched with \totalspe textual species descriptions and annotations for detection and key-part localization. We outline our comprehensive data collection methodology below, which includes a rigorous process for taxonomic correction, description generation, and the collection of bounding box and key-point annotations. To ensure the scientific validity of our benchmark, this entire process was conducted in close collaboration with experts in marine biology.

\begin{figure}[!t]
    \centering
    \includegraphics[width=1\linewidth]{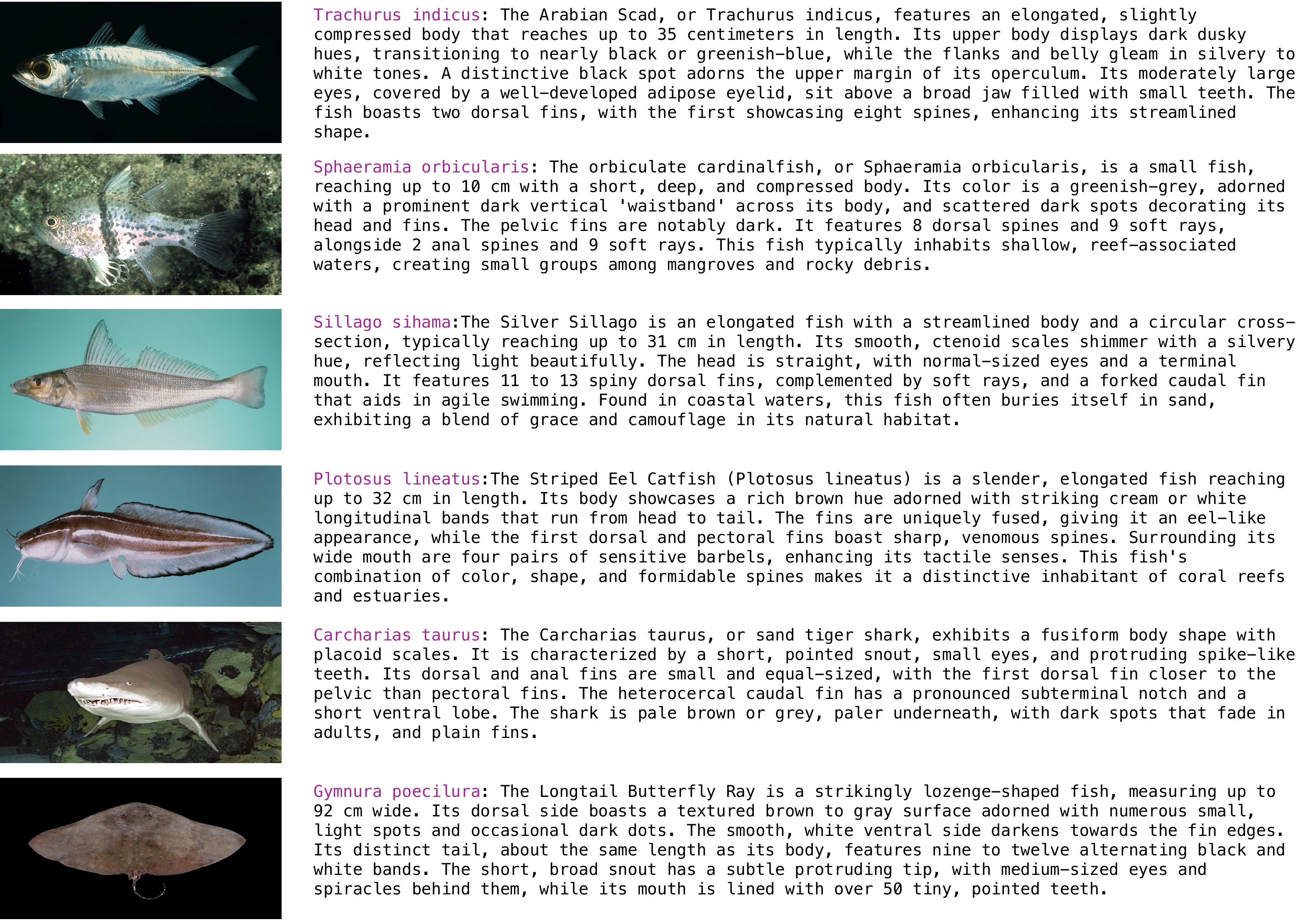}
    \caption{Examples of species description summarized by GPT-4o~\cite{gpt4o} using information scraped from credible sources as described in ~\cref{subsec:description}. }
    \label{fig:text}
\end{figure}

\subsection{Species Description} \label{subsec:description}
To generate descriptive text for each species, we first identified multiple reliable sources to serve as our knowledge base. FishBase~\cite{fishbase2025} was used as the primary source of morphological and ecological information, and supplemented by iNaturalist~\cite{inat_web}, WoRMS~\cite{worms2025}, and NOAA~\cite{noaa2025}. For species available on FishBase, $21,279$ out of the known $\totalspe$ fish species, we extracted detailed morphological data directly. For the remaining $13,854$ species missing morphological information from FishBase, we crawled iNaturalist, WoRMS, NOAA, and other supplemental sources to collect all available information. We then used GPT-4o~\cite{gpt4o} to consume the information and produce a coherent and concise descriptive summary of each species. To validate the reliability of the generated descriptions, a subset of fifty descriptions was examined by experts, confirmed to be of reasonable accuracy and to be visually discriminative within the constraints of the description parameters (i.e., coherent and concise).

We also evaluated the description in a user study. The users are shown four images of the corresponding species along with the description, and they are asked to rate the description on a 1-5 scale, with 1 indicating "not helpful at all" and 5 indicating "very helpful" for identifying the species. This was done for 1,000 marine species descriptions. Each description was rated by three human annotators. The descriptions received a mean score of 3.9, a median of 4.0, and a mode of 4.0, highlighting that the descriptions are of good quality for recognition.

\subsection{Key-Point Selection and Collection}
We finalized six-part locations and one attribute to be collected for every image in \ours. The parts are as follows: 1) Eye location, 2) Mouth location, 3) Pectoral, pelvic, and anal fin location, 4) Center of the main body, 5) Tail (caudal fin) start, and 6) Tail end. All the parts were annotated by pixel location in each image. Additionally, we record whether the species is underwater or above water. Fin locations may involve multiple points depending on the number of fins, with variations by species, and are subject to image angle and occlusion. Similarly, the apex of the tail can have one or two location points depending on the shape of the tail. The selection of these parts and attributes was done in consultation with experts to ensure the dataset's utility for both the machine learning and aquatic science communities. A key piece of information provided by the key points is the aspect ratio, which has been linked to species' behaviour, metabolism, ecological lifestyle, and response to thermal stress~\cite{Sambilay1990, CAMPOS2018148, anthropo}. This information can therefore be valuable in understanding species' ecology and can contribute to conservation decision-making. Additionally, key-part location can serve as weak supervision to obtain dense annotations like segmentation. Further discussed in~\cref{app:seg}.

To collect the part location annotations, we partnered with a company specializing in data annotations. Experts supervised and validated the annotation process to ensure quality control. Once the annotators were familiar with the process, we implemented a system of regular manual checks to maintain the quality of the part location annotations. \ours includes $86,589$ instances of eye locations, $77,990$ instances of mouth locations, $281,426$ instances of fin locations, $80,653$ instances of body locations, $73,785$ instances of tail-start locations, and $105,983$ instances of tail-end locations. In total, we provide $\kp$ key part locations for our dataset. From these images, $38,326$ images of fish are above the water surface. 

\begin{figure}[!t]
    \centering
    \includegraphics[width=1\linewidth]{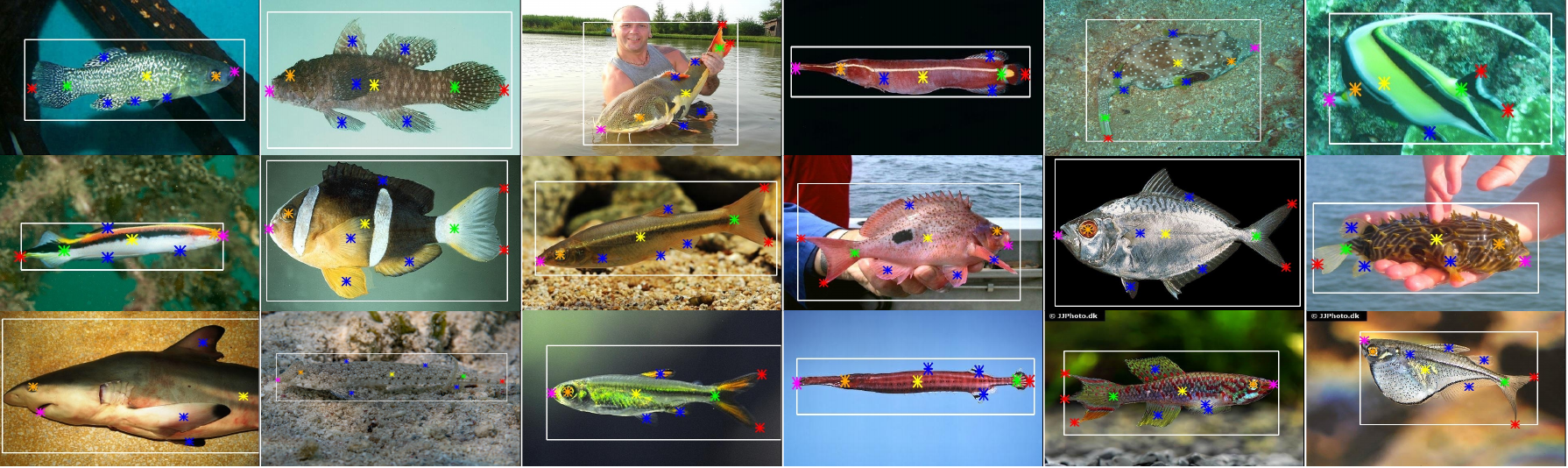}
    \caption{Example images from \ours showcasing part-level annotations. Each keypoint is color-coded by semantic part: eye (orange), fins (blue), mouth (magenta), body center (yellow), tail start (green), and tail apex (red). The number and placement of fins vary across species, and some species exhibit a forked tail apex. For each image, we also display the annotated bounding box.}
    \label{fig:data}
\end{figure}

\subsection{Taxonomic Corrections}
The taxonomy of species around the world is continuously evolving~\cite{new_species}, making it essential to ensure that datasets reflect the most up-to-date and accurate species names. During our analysis, we found that $266$ species names from the FishNet~\cite{fishnet} dataset no longer aligned with current taxonomic standards (as per~\cite{fishbase2025}). To address this, we manually remapped these outdated names to their correct, updated counterparts. Following this, we also associated each species in our dataset with its corresponding species code from FishBase, which remains the same even as taxonomic names change. This provides a straightforward mechanism for keeping our dataset aligned with current taxonomic nomenclature. In the end, we identified 36 images that did not correspond to any known species from the entire taxonomy. These may represent entirely new species to science.

\subsection{Additional Images}
The original Fishnet~\cite{fishnet} is highly long-tailed, with only \oldten species with ten or more images. The bias in the number of images is often associated with those species that are not exploited commercially at large scales (either for fisheries, ecotourism, or the aquarium trade), those that are found in less well-researched parts of the world, or those found in less accessible depth ranges.  For \ours, we sought to increase the number of species with reasonable image representation. We collected additional images for species from various underrepresented regions worldwide, including Egypt, Indonesia, Oman, Seychelles, Papua New Guinea, and Saudi Arabia, sourced through a wider network of collaborators who provided access to their extensive collections. In total, we gathered an additional \newcnt images, increasing the number of species with at least ten images from \oldten to \newten, significantly enhancing its diversity and representation.

\section{Experiments}
\label{sec:setup}

Based on \ours, we first evaluate the performance of various VLMs and MLLMs on the task of fish recognition. This is followed by a thorough analysis to explain the poor performance.

\subsection{Data Splits}
We follow a 75-5-20 train-validation-test split strategy for species with a sufficient number of images. Specifically, for species with at least 5 images, 75\% of the images are used for training, 5\% for validation, and the remaining 20\% for testing. For species with 3 or 4 images, we assign one image to the test set and use the remaining images for training. Species with fewer than 3 images (i.e., only 1 or 2 images) are not included in the main split. Instead, these rare cases are grouped into a separate "rare split", which exclusively contains species represented by 1 or 2 images. This splitting strategy is inspired by FishNet~\cite{fishnet}, which drops species with very few samples (1 or 2) for the classification experiments. However, in contrast, we retain these underrepresented species in the rare split to thoroughly evaluate the recognition capabilities of vision-language models. The test set contains $15,518$ images, while the rare set contains $16,367$ images. The frequent set consists of $5,584$ species, and the rare set consists of $11,809$ species.

\begin{wraptable}[16]{t}{0.5\textwidth} 
  \centering
  \vspace{-5mm}
\caption{Classification Accuracy: Evaluation of various open-source VLMs and MLLMs on the fish species open-vocabulary recognition task from species descriptions. Highest performance is in bold, and second-highest is in underline.}
  
\centering
\scalebox{0.75}{
\footnotesize
\begin{tabular}{l|rr}
Method & Frequent Species & Rare Species \\
\midrule
OpenCLIP~\cite{openclip} & 1.0 & 0.2\\
BioCLIP~\cite{bioclip} & 2.3 & 0.2  \\
CLIP~\cite{clip} & 2.4 & 0.2 \\
SigLIP~\cite{siglip} & 2.6 & \underline{0.5} \\
\midrule
LLaVa-Next~\cite{liu2023improved}  & 0.3 & 0.1\\
LLaVaOne~\cite{llavaone} & 0.6 & 0.0 \\
MiniCPM-V-2.6~\cite{minicpm} & 1.7 & 0.1 \\
InternVL-2.5~\cite{intern25} & 2.0 & 0.0 \\
Pixtral-12b~\cite{pixtral12b} & 3.6 & 0.1 \\
Gemma-3~\cite{gemma3} & 5.5 & 0.2 \\
Qwen2.5-VL~\cite{qwen25} & \underline{6.2} & 0.2 \\
GPT-4o & \textbf{17.9} & \textbf{1.2} \\
\midrule
\end{tabular}
}

\label{tab:baseline}
\end{wraptable}

\subsection{Recognition Results}
\label{sub_ovc}
Unlike traditional classification tasks that rely on a closed and predefined label space~\cite{survey}, this task operates under an open and continually expanding set of species labels. To address this challenge, we leverage Vision-Language Models (VLMs) and MLLMs while utilizing all $35,133$ textual descriptions of species to infer the species present in the image. For CLIP-based VLMs~\cite{clip, openclip, siglip}, the approach is straightforward: we compute the cosine similarity between the visual embedding of an input image and the text embeddings of species descriptions. When species descriptions exceed the model's context length, we chunk them appropriately. The species whose description yields the highest similarity is selected as the predicted label. To evaluate MLLMs, we formulate the task as a ``Question Answering" task, where the question is to identify the species present in the image. We compare CLIP~\cite{clip}, OpenClip~\cite{openclip}, BioClip~\cite{bioclip}, and SigLip~\cite{siglip} as our VLM baselines. For MLLMs, we include InternVL-2.5(8B)~\cite{intern25}, MiniCPM(8.1B)~\cite{minicpm}, Gemma-3(12.2B)~\cite{gemma3}, Pixtral-12b(12B)~\cite{pixtral12b}, LlaVa-Next(13.4B)~\cite{liu2023improved}, LlaVaOne(8.03B)~\cite{llama}, and Qwen2.5-VL(8.29B)~\cite{qwen25}. We also include GPT-4o~\cite{gpt4o} as a representative closed-source model.

As shown in~\cref{tab:baseline}, all models face significant challenges in accurately recognizing fish species from images, highlighting the difficulty of fine-grained open-world classification in the marine domain. Among all open-source models, Qwen2.5-VL achieves the highest performance on frequent species, followed by Gemma-3, while SigLIP performs best on the rare species subset. Although the overall accuracy remains low, it is still three orders of magnitude better than random guessing, highlighting the models' ability to learn some meaningful signal despite the task's difficulty.

\subsubsection{Results at Genus Level}
The species-level results indicate that current models do not yet achieve a strong overall performance, highlighting the difficulty of fine-grained, open-vocabulary classification. To investigate whether this challenge is alleviated at coarser taxonomic levels, we analyze whether the species predicted by the models belong to the correct genus. 
\begin{wraptable}[14]{t}
{0.5\textwidth} 
  \centering
  \vspace{-2mm}
\caption{Classification Results at the Genus Level. Highest performance is in bold, and second-highest is in underline.}
\vspace{-2mm}
  \centering
\scalebox{0.7}{
\footnotesize
\begin{tabular}{l|rr}
Method & Frequent Species & Rare Species \\
\midrule
OpenCLIP~\cite{openclip} & 5.0 & 2.3 \\
BioCLIP~\cite{bioclip} & 8.5 & 3.2  \\
CLIP~\cite{clip} & 9.4 & 3.8 \\
SigLIP~\cite{siglip} & 14.8 & \underline{8.6} \\
\midrule
LLaVa-Next~\cite{liu2023improved}  & 5.6 & 0.7 \\
LLaVaOne~\cite{llavaone} & 2.5 & 0.6 \\
MiniCPM-V-2.6~\cite{minicpm} & 6.0 & 1.3 \\
InternVL-2.5~\cite{intern25} & 6.8 & 0.7 \\
Pixtral-12b~\cite{pixtral12b} & 8.2 & 3.0 \\
Gemma-3~\cite{gemma3} & 14.3 & 3.0 \\
Qwen2.5-VL~\cite{qwen25} & \underline{18.2} & 5.1 \\
GPT-4o & \textbf{34.2} & \textbf{14.1} \\
\midrule
\end{tabular}
}

\label{tab:baseline_genus}
\end{wraptable}
In~\cref{tab:baseline_genus}, we report the genus accuracy for all the models. 
We calculate the genus accuracy by mapping all species-level predictions made by each model to their corresponding genus. This allows us to distinguish between fine-grained misclassifications within the same genus and truly incorrect predictions across unrelated taxa. Compared to species-level results, we observe a clear improvement in accuracy, indicating that while models struggle with the extreme fine-grained species classification, they often predict the correct genus. We further extend this analysis to the family-level taxonomy in ~\cref{app:res}. The performance improves substantially at the Family level, with Qwen2.5-VL and GPT-4o achieving 30.5\% and 53.6\% accuracy for frequent species, and 14.3\% and 37.4\% for rare species, respectively.

\begin{wraptable}[7]{t}{0.5\textwidth} 
  \centering
\caption{Performance of MLLMs on bidirectional name task.}
\scalebox{0.80}{
\footnotesize
\begin{tabular}{lcc}
\toprule
Method & Common $\rightarrow$ Scientific & Scientific $\rightarrow$ Common\\
\midrule
Qwen2.5-VL & 3.6 & 3.6 \\
\bottomrule
\end{tabular}
}
\label{tab:kdg}
\end{wraptable}

\subsection{Domain Knowledge}
To investigate whether the poor performance of MLLMs stems from a foundational knowledge gap. We devised a bidirectional name translation task using the common and scientific names for all \totalspe species, sourced from FishBase~\cite{fishbase2025}. We evaluated the top-performing open source MLLM (Qwen2.5-VL) on its ability to map a scientific name to any of its corresponding common names, and conversely, a common name to its single correct scientific name. As shown in Table~\ref{tab:kdg}, the model struggles significantly with this task for marine species, with a mere 3.6\% correct translations. In stark contrast, the same evaluation performed on the CUB-200-2011 bird dataset~\cite{cub} yielded an accuracy of 40.0\%. This discrepancy strongly suggests that the model's failure is not a general limitation but lacks the basic taxonomic information needed to link common and scientific names, a task that requires no visual understanding.

\begin{wraptable}[6]{t}{0.5\textwidth} 
  \centering
  \vspace{-8mm}
\caption{Confusion matrix for the fine-grained differentiation task}
\scalebox{0.5}{
\footnotesize
\begin{tabular}{lcccc}
\toprule
& \multicolumn{2}{c}{\textbf{Correct Species (Positive Case)}} & \multicolumn{2}{c}{\textbf{Incorrect Species (Negative Case)}} \\
\cmidrule(lr){2-3} \cmidrule(lr){4-5}
\textbf{Method} & \textbf{TP Rate (\%)} & \textbf{FN Rate (\%)} & \textbf{TN Rate (\%)} & \textbf{FP Rate (\%)} \\
\midrule
Qwen2.5-VL(random) & 81.4 & 18.6 & 67.1 & 32.9 \\
Qwen2.5-VL(fine-grain) & 56.4 & 43.6 & 34.8 & 65.2 \\
Random Chance & 50.0 & 50.0 & 50.0 & 50.0 \\
\bottomrule
\end{tabular}

}
\label{tab:fgd}
\end{wraptable}

\subsection{Visual Domain Knowledge}
Having established the MLLM's semantic knowledge deficit with the name translation task, we next investigated if this was compounded by a failure in visual perception. For this, we designed a species verification task where the model was given an image and asked if a candidate species was present in the image or not. The task was repeated twice, once with the correct candidate and once with the wrong candidate. The wrong candidate was chosen either at random or was chosen from the nearest neighbors of the correct candidate in the CLIP space.

From~\cref{tab:kdg}, it is clear that Qwen2.5-VL can distinguish if the species is present or not when the candidate is chosen at random, but when the candidate is more fine-grained, the model mostly answers 'Yes'. The average performance of Qwen2.5-VL for the fine-grained case is slightly worse than random chance. The poor performance on the fine-grained case reveals a failure in visual domain knowledge. Its inability to reliably accept the correct species and, crucially, reject the visually similar incorrect species, demonstrates that the issue is twofold. It not only lacks the deep domain knowledge to understand the subtle differences between species but also the fine-grained perceptual ability to discern those differences in an image. This shows the model's knowledge gap is not purely abstract but is also related to its visual processing capabilities. However, this test does not distinguish between coarse and fine-grained perception. To investigate this, we next evaluate the models on object detection and key-part localization.

\begin{table}[t]
\caption{Performance of Qwen2.5-VL on coarse-grained detection. The performance is reported using IoU thresholds (50–90).}
\small
\begin{center}
\scalebox{0.8}{
\begin{tabular}{l|cccccccccccc}
\toprule
\multirow{2}{*}{Method} & \multicolumn{5}{c}{Frequent Species (\%)} & & \multicolumn{5}{c}{Rare Species (\%)} \\
\cline{2-6} \cline{8-12}
 & IoU50 & IoU60 & IoU70 & IoU80 & IoU90 & & IoU50 & IoU60 & IoU70 & IoU80 & IoU90 \\
\midrule
YOLO-12~\cite{yolov12} & \textbf{95.2} & \textbf{92.0} & \textbf{84.7} & \textbf{67.9} & \textbf{35.2} && \textbf{95.6} & \textbf{93.3} &	\textbf{88.1} & \textbf{74.1} & \textbf{40.6} \\
Qwen2.5-VL~\cite{qwen25} & 91.5	& 85.1 & 73.8 & 54.8 & 26.7 & & 95.2 & 91.0 & 82.0 &	63.8 & 31.3\\
\bottomrule
\end{tabular}
}
\end{center}

\label{tab_detection}
\end{table}

\subsection{Perception Capabilities}
\noindent\textbf{Coarse-Grained Vision}: Before fine-grained recognition, a model must first perform coarse-grained visual localization, that is, correctly identifying the object's location within an image. Failure at this initial stage makes recognition unlikely. To assess this capability, we evaluated Qwen-VL on a detection task, using the ground-truth bounding box coordinates from \ours. The task is relatively straightforward, as our dataset predominantly contains single-instance images.

In~\cref{tab_detection}, we compare Qwen2.5-VL with YOLO-12~\cite{yolov12}, trained on \ours. While Qwen2.5-VL underperforms YOLO-12, its results show a strong ability to localize fish, suggesting that recognition failures stem less from object detection and more from knowledge gaps or limitations in fine-grained visual perception, which we investigate next.

\begin{table}[t]
\caption{Performance of Qwen2.5-VL and a YOLO-based baseline on the fine-grained vision task. }
\scalebox{0.7}{
\footnotesize
\begin{tabular}{l|ccccccccccccc}
\toprule
\multirow{2}{*}{Method} & \multicolumn{6}{c}{Frequent Species (\%)} & & \multicolumn{6}{c}{Rare Species (\%)} \\
\cline{2-7} \cline{9-14}
 & Tail End	& Fin & Tail Start & Body & Mouth & Eye && Tail End	& Fin & Tail Start & Body & Mouth & Eye \\
\midrule
YOLO-12~\cite{yolov12} & \textbf{30.8} & \textbf{16.6} & \textbf{46.6} & \textbf{45.9} & \textbf{45.7} & \textbf{44.7} &&   \textbf{29.7} & 15.2 & \textbf{46.1} & \textbf{46.4} & \textbf{45.8} & \textbf{43.3} \\
Qwen2.5-VL~\cite{qwen25} & 23.4 & 15.6 & 21.8 & 36.8 & 27.5 & 27.1 && 26.1 & \textbf{16.6} & 22.3 & 37.4 & 26.4 & 27.2 \\
\bottomrule
\end{tabular}
}

\label{tab_kp}
\end{table}

\noindent\textbf{Fine-Grained Vision}: 
To test the fine-grained visual capabilities of the Qwen model, we evaluate it on the task of key-part localization, where the model is required to identify the precise locations of body parts. In~\cref{tab_kp}, we report the PCK~\cite{pck1, pck2} score, which measures the proportion of keypoints that lie within a certain distance from the ground truth relative to the object size. The results in~\cref{tab_kp} compare the Qwen model against a YOLO model trained on the \ours training set. Unlike the detection task, Qwen performs significantly worse on key-part localization. This poor performance highlights a critical limitation, as many fish species closer in the taxonomic tree often differ only in subtle, fine-grained details, such as eye coloration, the exact number of fin rays, or the shape of the tail. Capturing and reasoning over such nuanced visual cues is essential for reliable species-level recognition. This experiment shows us that while MLLMs like Qwen demonstrate competence on coarse-level recognition, they lack the fine-grained perceptual sensitivity needed for domains such as biodiversity research.

\subsection{Does Finetuning help?}

To enhance the zero-shot recognition performance of Qwen2.5-VL, we explore multiple approaches.

\begin{figure}[!t]
    \centering
    \includegraphics[width=1\linewidth]{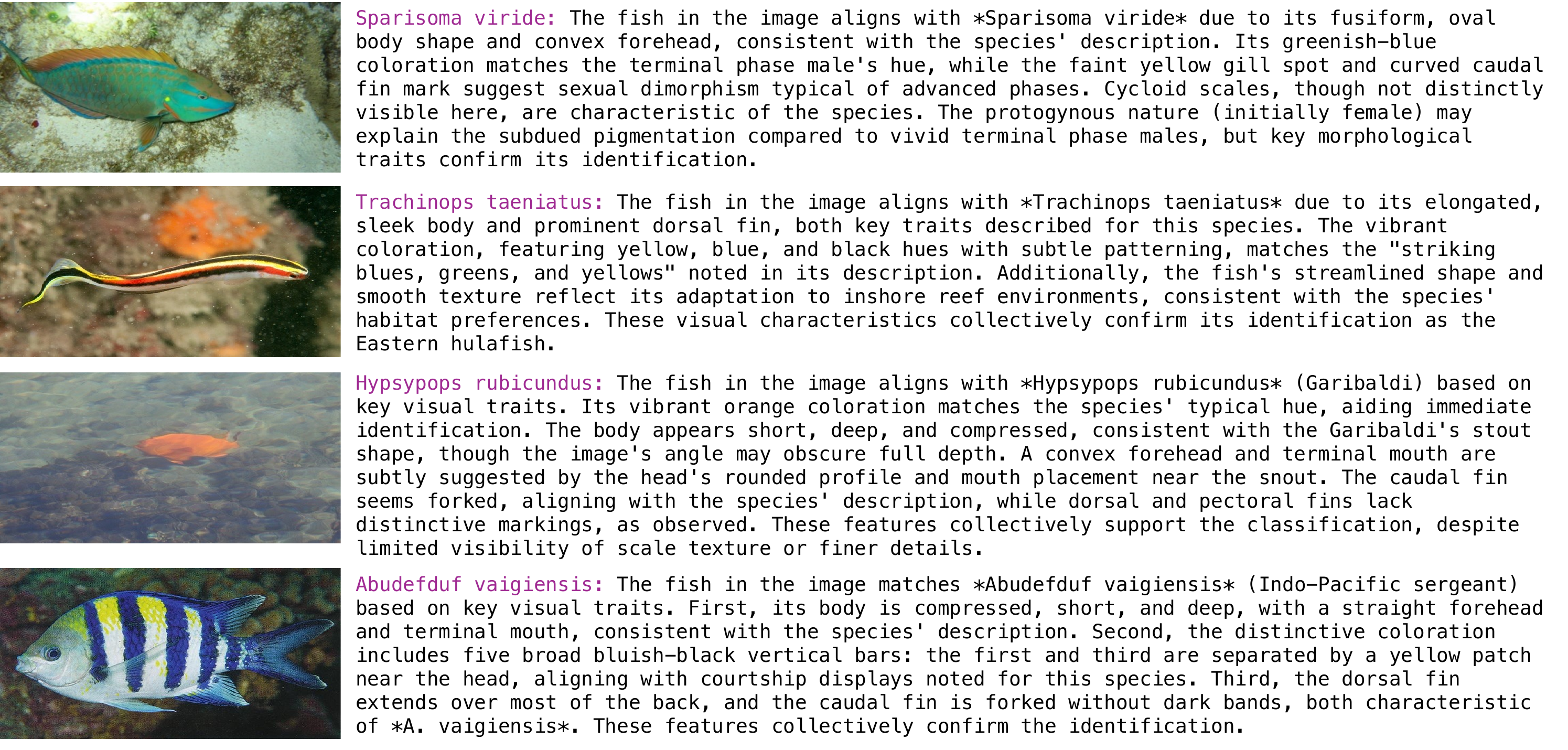}
    \caption{Qualitative examples of species identification and reasoning generated by our finetuned Qwen-VL model when trained for explainability.}
    \label{fig:qual_int}
\end{figure}

\noindent 1) \textbf{retrieval-augmented generation}: The model is provided with textual context from species descriptions, but RAG performance depends heavily on the retrieval step. Since this is a cross-modal task, VLMs like CLIP are natural candidates, yet they perform poorly (see~\cref{tab:baseline}). To improve retrieval, we use an ensemble of CLIP, BioCLIP, and SigLIP, which yields more accurate candidates. The top 10 retrieved species descriptions are then passed to Qwen2.5-VL as context. This ensemble-based RAG (E-RAG) improves performance by $\approx$1\% (see~\cref{tab:sft}), with potential for further gains using stronger retrievers. Retrieval results for individual VLMs are detailed in~\cref{app:ret}.

\noindent 2) \textbf{supervised-finetuning}: To enhance recognition performance, we performed LoRA-based supervised finetuning (SFT) on the Qwen-VL model using our proposed training set. This process substantially improved accuracy on frequent species, from a zero-shot baseline of 6.2\% to 37.0\%. We compare this against two strong baselines: pre-trained ViT~\cite{dosovitskiy2021an} and BEiT~\cite{beit3} finetuned on \ours. While our finetuned Qwen-VL outperforms the ViT baseline, BEiT achieves the highest accuracy at the species level. However, an analysis at higher taxonomic levels reveals that the finetuned Qwen-VL surpasses BEiT. This suggests that while BEiT may overfit to specific species-level features, Qwen-VL learns a more semantically robust representation, producing predictions that are taxonomically closer to the ground truth. 

\begin{wraptable}[12]{t}{0.4\textwidth} 
\vspace{-5mm}
  \centering
\caption{Finetuned classification results. Qwen2.5-VL ft. represents a finetuned version, and int. represents a finetuned version with reasoning.}
  \centering
\scalebox{0.75}{
\footnotesize
\begin{tabular}{l|rrr}
\toprule
\multirow{2}{*}{Method} & \multicolumn{2}{r}{Accuracy} \\
 & Species & Genus & Family\\
\midrule
ViT & 25.3 & 31.5 & 38.4\\
BeiT & \textbf{43.4} & 50.9 & 58.2\\
\midrule
Qwen2.5-VL & 6.2 & 18.2 & 30.5\\
Qwen2.5-VL + RAG & 4.8 & 15.7 & 21.6\\
Qwen 2.5VL + E-RAG & 7.1 & 22.7 & 46.2\\
Qwen2.5-VL ft. & \underline{37.0} & \textbf{51.5} & \underline{64.7}\\
Qwen2.5-VL int. & 35.4 & 51.0 & \textbf{65.4}\\
\midrule
\end{tabular}
}

\label{tab:sft}
\end{wraptable}

\noindent 3) \textbf{Explainable supervised-finetuning}: To fully leverage \ours, we finetune Qwen to predict the correct species and generate supporting reasoning. This auxiliary task incurs minimal cost to recognition performance while greatly improving interpretability.

To construct the training corpus of reasoning, we employ GPT-4.0, which is provided with the input image(from the training set), the candidate species, and the species description, and asked to generate a concise justification for why the image corresponds to the given species. These reasoning texts are then paired with the species labels and used jointly during finetuning.  We report the overall performance of Qwen under different training settings in~\cref {tab:sft}, and additionally provide qualitative examples showcasing both predictions and their associated reasoning in~\cref{fig:qual_int}. Beyond accuracy, the generated explanations make the model's decisions more transparent and interpretable. Such interpretability is particularly valuable for marine scientists, as it enables verification of the model's decision-making process, facilitates error analysis when misclassifications occur, and provides human-readable insights that can support downstream ecological studies.

\section{Conclusion}
\label{sec:conclusion}

In this work, 
\begin{enumerate}
    \item We introduce \ours, a comprehensive multimodal benchmark for marine species recognition, designed to evaluate the strengths and limitations of MLLMs on fine-grained ecological tasks, offering textual descriptions, bounding boxes, and key-part annotations.
    \item Our analysis reveals that state-of-the-art VLMs and MLLMs struggle with fine-grained taxonomic and morphological distinctions despite general recognition ability.
    \item Through diagnostic experiments, we disentangle errors from domain knowledge gaps, weak visual perception, and limited reasoning.
    \item Fine-tuning on \ours narrows the performance gap, and explainable fine-tuning further boosts interpretability, underscoring the importance of domain-specific benchmarks.
\end{enumerate}  

\bibliography{iclr2025_conference}
\bibliographystyle{iclr2025_conference}

\appendix
\clearpage

\section{Experimental Details.} 
\label{app:exp}
All inferences were performed on a single A100 GPU. For VLMs, the species prediction was made by selecting the class belonging to the chunk with the highest similarity score. For MLLMs, the species name was generated via prompting. We used two prompt variants:  1) without context: \textit{Given the image of the fish, please answer with the species to which the fish belongs to? Only answer with the species scientific name.} and with RAG: \textit{I have an image of a fish and need to identify its species type. I have narrowed it down to ten possible species. Please use the following descriptions to determine the most likely species: \{\}. Analyze the fish in the image, considering its physical characteristics, and compare them to the given species descriptions. Provide only the name of the most likely species.} In the RAG setting, we provided the MLLMs with the top 10 retrieved species descriptions. We also conduct an ablation study varying the number of descriptions fed to the MLLMs. 

\noindent{\textbf{Training Details}}
LORA-based supervised fine-tuning of Qwen was done on 4-A100 GPUs with 80GB memory for 4000 steps, with an effective batch size of 32, rank 8. Optimization was conducted using AdamW, employing an initial learning rate of 0.0001 with a cosine learning schedule and a 0.1 warmup ratio.

YOLO-based model training was performed using the Ultralytics YOLO framework for 30 epochs with a mini-batch size of 16 images. All input images were uniformly resized to 640×640 pixels. Optimization was conducted using Stochastic Gradient Descent on an NVIDIA V100 GPU with 32 GB of memory, employing an initial learning rate of 0.01, a momentum factor of 0.937, and a weight decay of 0.0005. A 3-epoch warmup phase was employed, linearly increasing the momentum from 0.8 and the bias learning rate from 0.1. We used the corresponding YOLO model as the base model with pre-trained weights utilized to speed up the convergence and enhance performance.

\begin{wraptable}[16]{t}{0.5\textwidth} 
  \centering
  \vspace{-5mm}
\caption{Classification Accuracy: Evaluation of various open-source VLMs and MLLMs on the fish family open-vocabulary recognition task from species descriptions. Highest performance is in bold, and second-highest is in underline.}
  \centering
\scalebox{0.75}{
\footnotesize
\begin{tabular}{l|rr}
& Frequent Species & Rare Species \\
\midrule
OpenCLIP~\cite{openclip} & 14.4 & 10.3 \\
BioCLIP~\cite{bioclip} & 17.7 & 12.7  \\
CLIP~\cite{clip} & 22.7 & 15.8 \\
SigLIP~\cite{siglip} & \underline{32.9} & \underline{28.8} \\
\midrule
LLaVa-Next~\cite{liu2023improved}  & 8.9 & 2.1 \\
LLaVaOne~\cite{llavaone} & 6.7 & 3.4 \\
MiniCPM-V-2.6~\cite{minicpm} & 13.1 & 5.4 \\
InternVL-2.5~\cite{intern25} & 12.9 & 2.9 \\
Pixtral-12b~\cite{pixtral12b} & 14.7 & 12.6 \\
Gemma-3~\cite{gemma3} & 24.6 & 12.5 \\
Qwen2.5-VL~\cite{qwen25} & 30.5 & 14.3 \\
GPT-4o & \textbf{53.6} & \textbf{37.4} \\
\midrule
\end{tabular}
}

\label{tab:family}
\end{wraptable}

\section{Higher Taxonomy Results.} 
\label{app:res}
We extend our evaluation to the family-level classification, building upon the species and genus-level results presented in the main paper. From Table~\ref{tab:family}, we can see that as we go higher in the taxonomic hierarchy, from species to genus to family, the classification task becomes less granular, leading to improved performance across models. This trend is consistent with the inherent structure of biological taxonomy, where higher-level categories encompass broader groupings of organisms. Notably, the relative performance of models remains consistent across taxonomic levels. Qwen2.5-VL continues to outperform other open-source models, and its performance is further enhanced through the integration of the Ensemble RAG framework.

\section{VLM Retrieval Performance.} 
\label{app:ret}
We report retrieval performance across both frequent and rare classes using Mean Reciprocal Rank (MRR) at 1, 5, and 10 in Table~\ref{tab:ret}. Among individual models, SigLIP consistently performs the best, achieving an MRR@10 of 4.5 on seen classes and 1.2 on unseen classes. In contrast, BioCLIP, CLIP, and OpenCLIP show lower performance individually, with OpenCLIP performing the worst overall with MRR@10 of 1.6 and 0.4 on seen and unseen classes, respectively.

The best retrieval performance is observed when we combine all three models, CLIP + OpenCLIP + BioCLIP, achieving an MRR@10 of 8.4 on seen classes and 1.2 on unseen classes. This demonstrates that model ensembling can significantly boost retrieval quality, particularly for seen species. However, across all models, performance drops substantially on unseen classes, highlighting the challenge of generalization in open-world species retrieval.
\begin{table}
  \centering
  \vspace{-5mm}
\caption{Mean Reciprocal Rank (MRR) at 1, 5, and 10 for retrieval performance on frequent and rare species. While individual models like SigLIP outperform others, the combination of CLIP, OpenCLIP, and BioCLIP yields the highest performance on seen classes. All models show a noticeable drop in performance on unseen species, clearly demonstrating the difficulty of generalization. Highest performance is in bold, and second-highest is in underline.}
  \centering
\setlength\tabcolsep{12pt}
\scalebox{0.75}{
\begin{tabular}{lcccccc}
\toprule
 & \multicolumn{3}{c}{\textbf{Frequent}} & \multicolumn{3}{c}{\textbf{Rare}} \\
 & MRR@1 & MRR@5 & MRR@10  & MRR@1 & MRR@5 & MRR@10\\
\cmidrule(r){1-1} 
\cmidrule(lr){2-4} \cmidrule(lr){5-7} 
BioCLIP~\cite{bioclip} & 2.3 & 3.3 & 3.5 & 0.2 & 0.4 & 0.4 \\
CLIP~\cite{clip} & 2.4 & 3.5 & 3.8 & 0.2 & 0.4 & 0.4 \\
OpenCLIP~\cite{openclip}& 1.0 &	1.5 & 1.6 & 0.2 & 0.3 & 0.4\\
SigLIP~\cite{siglip} & \underline{2.6} & \underline{4.1} & \underline{4.5} & \underline{0.5} & \textbf{1.0} & \textbf{1.2} \\
\textbf{E-RAG} & \textbf{5.5} & \textbf{7.9} & \textbf{8.4} & \textbf{0.6} & \textbf{1.0} & \textbf{1.2} \\
\bottomrule
\end{tabular}
}

\label{tab:ret}
\end{table}

\section{Segmentation}
\label{app:seg}
\begin{figure*}
    \centering
    \includegraphics[width=1\linewidth]{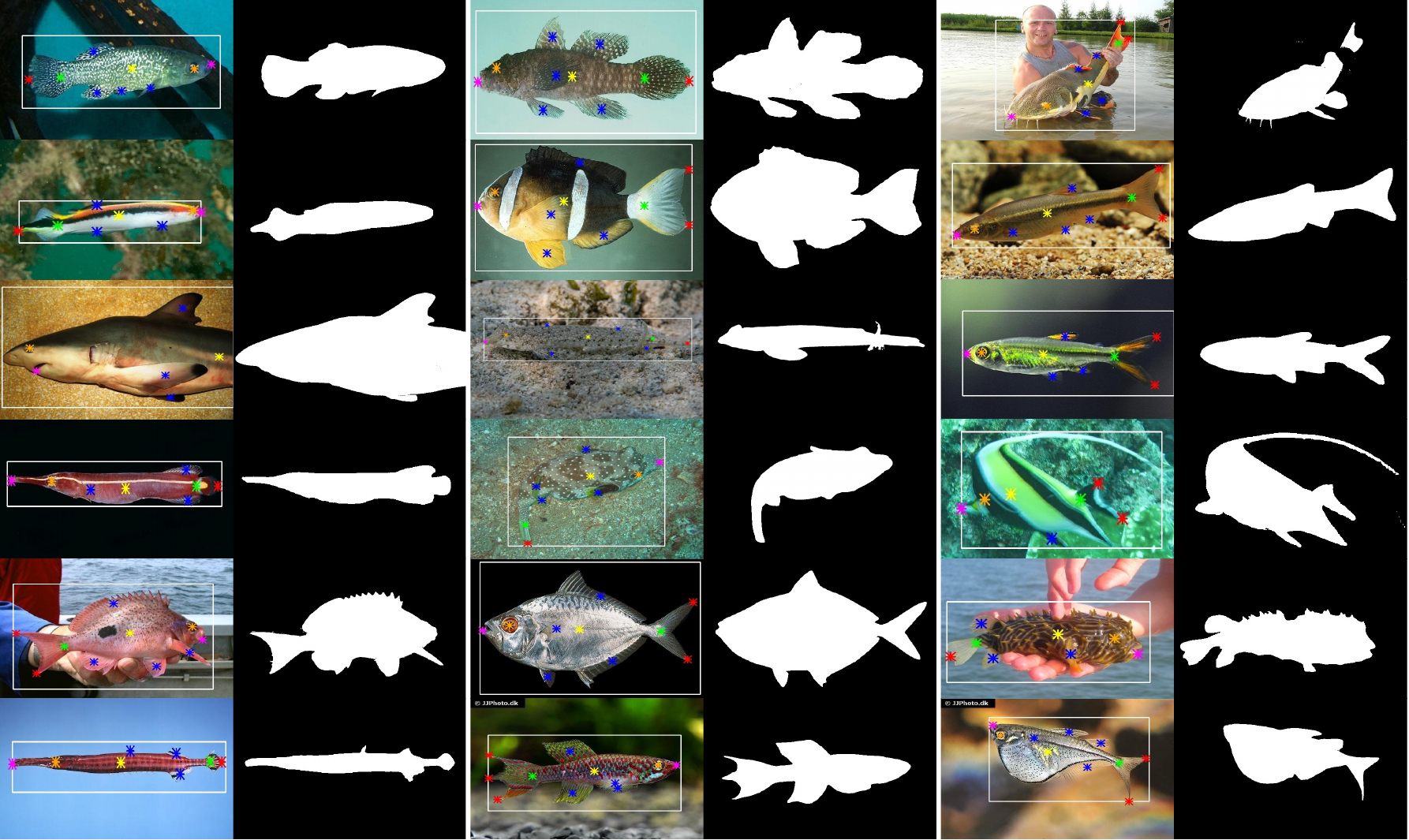}
    \caption{We show the same images from~\cref{fig:data} with segmentation masks obtained from our automated pipeline using key-points as supervision.}
    \label{fig:data_supp}
\end{figure*}

Obtaining segmentation masks is often a time-consuming and labor-intensive task. To address this, we employed a semi-automated pipeline to generate segmentation masks for \ours. Specifically, we used keypoints collected for each image as prompts to guide the Segment Anything Model~\cite{sam}, enabling it to better infer the approximate structure of the target object. This keypoint-guided approach proved highly effective. Some examples are shown in~\cref{fig:data_supp}. 

\begin{figure*}
    \centering
    \includegraphics[width=1\linewidth]{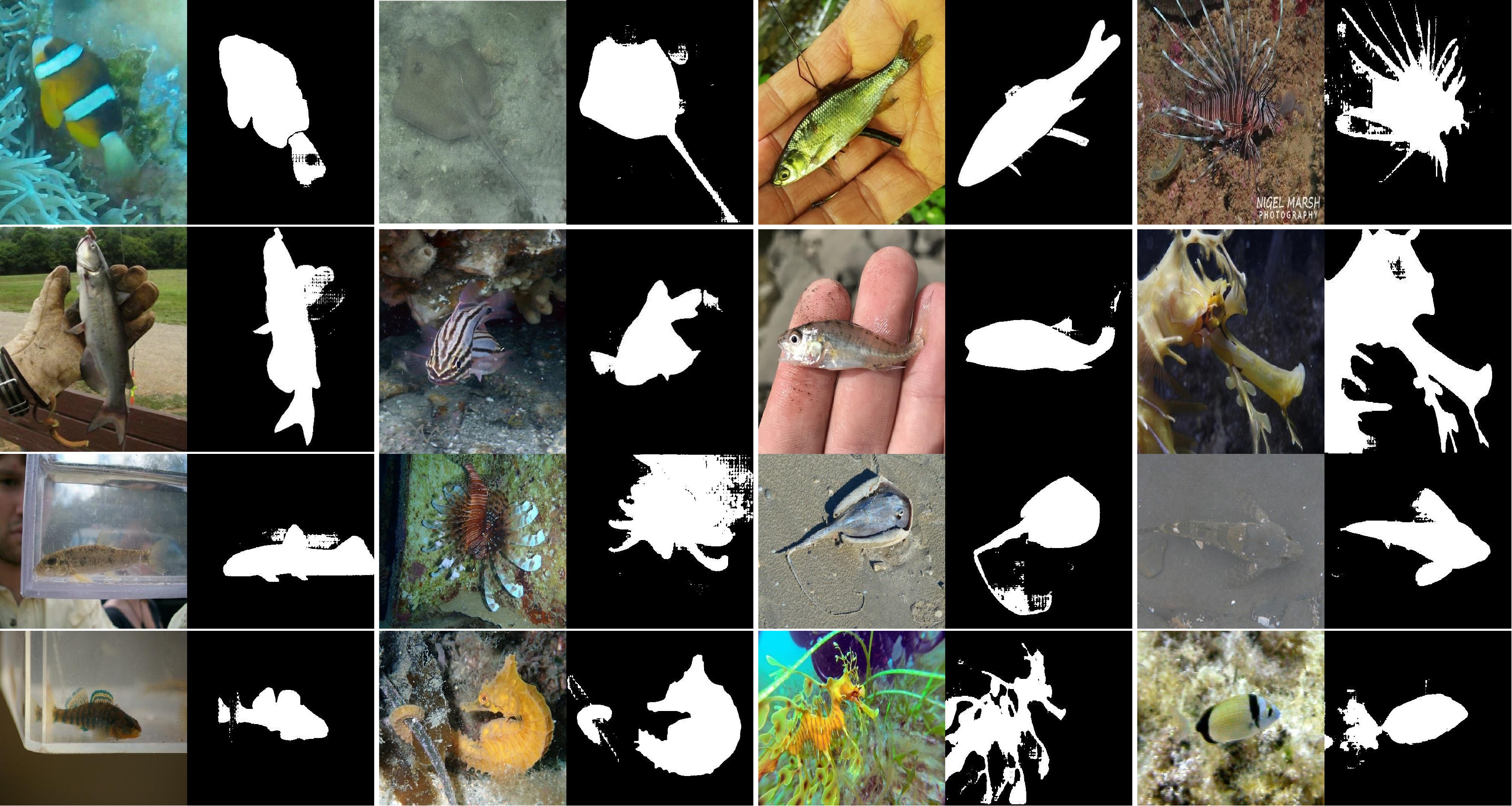}
    \caption{We show the samples with erroneous segmentation masks obtained from our automated pipeline using key-points as supervision.}
    \label{fig:wrong_seg}
\end{figure*}

To evaluate the effectiveness of this approach, we performed a manual evaluation of the generated segmentation mask from both the frequent and rare species test sets. In this test, the annotators were asked if the given segmentation mask completely covered the marine species without missing any part of its body. From 31,885 images of the two test sets, 24,278(76\%) were considered to be perfect by users, and the remaining ones captured most of the body but often missed parts like tails and fins, as shown in~\cref{fig:wrong_seg}. This clearly shows that our approach is highly effective for obtaining automated segmentation masks.

\section{Crowdsourcing Details.} 
\label{app:data}
\begin{figure}[t]
  \centering
  \includegraphics[width=\linewidth]{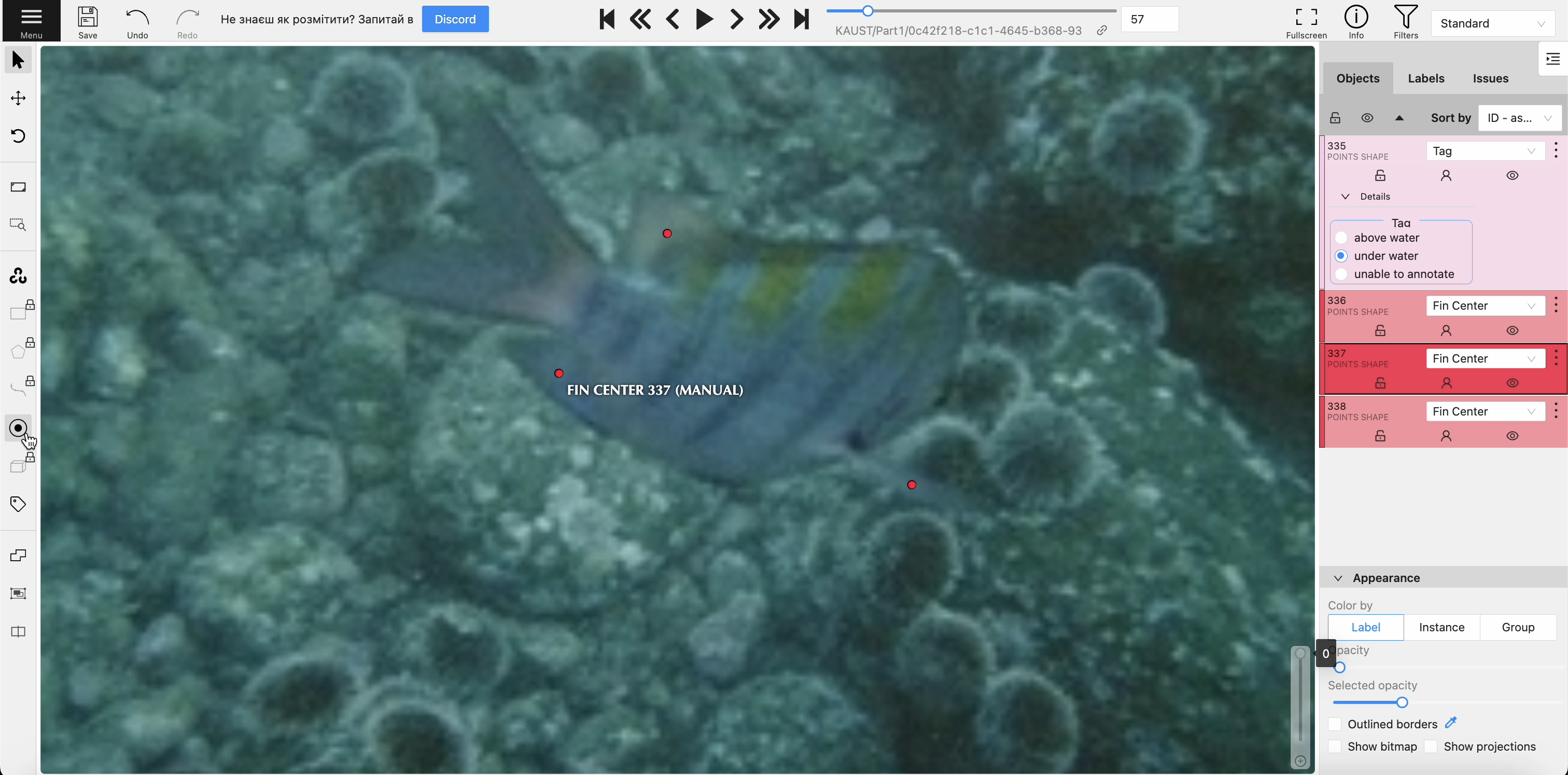}
  \caption{This figure illustrates the interface used for part location annotation. Here, the user has zoomed into the image to accurately label the center of the fin. Two fins have already been labeled, as indicated by the red-colored dot at the fin centers. The entire labeling process is efficient and user-friendly, as demonstrated in the video clip available \href{https://drive.google.com/file/d/1f36YdDNESjW_mb5scdreYtw2XHiet_1a/view?usp=sharing}{here}}.
  \label{fig:lyd}
\end{figure}

To enable efficient and accurate collection of data, we worked with an annotation service provider~\footnote{\href{https://labelyourdata.com/}{https://labelyourdata.com/}}. The custom-designed interface was developed to facilitate the collection and verification of part location and segmentation masks. We show the interface in Figure~\ref{fig:lyd} and also include~\href{https://drive.google.com/file/d/1f36YdDNESjW_mb5scdreYtw2XHiet_1a/view?usp=sharing}{link to a video clip} to completely demonstrate the annotation process.

\section{Limitations.} 
\label{app:lim}
Despite the extensive coverage and high-quality annotations provided by FishNet++, several limitations remain:
\begin{itemize}
    \item While FishNet++ includes a large number of species and diverse annotations, the dataset is still constrained by available imagery. Certain ecological regions and rare species remain underrepresented, limiting the generalizability of models trained on this data to truly global scenarios that contain over 35,000 species.
    \item Prompt-based evaluation for MLLMs can be highly sensitive to the structure and content of the prompt, which may introduce bias in comparisons. Further, large models may hallucinate plausible but incorrect species names, particularly under open-vocabulary settings.
    \item Underwater imagery presents extreme domain shifts (lighting, turbidity, occlusion) that remain difficult for both MLLMs and task-specific models. Performance in these conditions, while informative, may not fully reflect real-time field performance.
\end{itemize}

\end{document}